\documentclass[twoside,leqno,twocolumn]{article}

\usepackage[letterpaper]{geometry}

\usepackage{ltexpprt}
\usepackage{hyperref}
\usepackage[numbers]{natbib}

\usepackage{fancyhdr}

\usepackage{algorithm}
\usepackage[noend]{algpseudocode}

\usepackage{amsfonts}
\usepackage{amsmath}
\usepackage{graphicx}
\usepackage{accents}
\usepackage{hyperref}
\usepackage{bbm}
\usepackage{bm}
\usepackage{booktabs}
\usepackage{float}
\usepackage{multirow}
\usepackage{scalerel}
\usepackage{bm}

\usepackage{tikz}
\usetikzlibrary{calc, decorations.pathmorphing}
\usetikzlibrary{decorations.pathreplacing, chains}
\usetikzlibrary {arrows.meta}

\newcommand{\RR}{\mathbb{R}}

\newcommand{\paren}[1]{\left( #1 \right)}
\newcommand{\br}[1]{\left[ #1 \right]}
\newcommand{\abs}[1]{\left| #1 \right|}

\newcommand{\norm}[1]{\left\lVert#1\right\rVert}

\newcommand{\scr}[1]{\mathcal{#1}}

\newcommand{\pluseq}{\mathrel{{+}}=}

\mathchardef\mhyphen="2D

\newcommand{\tp}{\textrm{TP}}

\usepackage{url}

\begin{document}

\newcommand\relatedversion{}

\title{\Large An Efficient Sparse Kernel Generator for O(3)-Equivariant
Deep Networks} 

\author{Vivek Bharadwaj\thanks{Equal Contribution} \thanks{University of California, Berkeley} \thanks{Lawrence Berkeley National Laboratory} \and Austin Glover\footnotemark[1] \footnotemark[2] \and Ayd{\i}n Bulu\c{c}\footnotemark[3] \footnotemark[2] \and James Demmel\footnotemark[2]}


\maketitle


\fancyfoot[R]{\scriptsize{Copyright \textcopyright\ 2025 by SIAM\\
Unauthorized reproduction of this article is prohibited}}





\begin{abstract} \small \baselineskip=9pt
Rotation equivariant graph neural networks, i.e.
networks designed to guarantee certain geometric relations between their inputs and outputs,
yield 
state of the art performance on spatial deep learning tasks. They
exhibit high data efficiency during training and
significantly reduced inference time for 
interatomic potential calculations compared to
classical approaches. Key to these models is the
Clebsch-Gordon (CG) tensor product, a kernel that
contracts two dense feature vectors with a 
highly-structured sparse tensor to produce a
dense output vector. The operation, which may
be repeated millions of times for typical equivariant
models, is a costly and inefficient bottleneck. We introduce a 
GPU sparse kernel generator for the CG
tensor product that provides significant
speedups over the best existing open 
and closed-source implementations. Our
implementation achieves high 
performance by carefully 
managing the limited GPU shared memory
through static analysis at model compile-time,
minimizing reads and writes to global memory.
We break the tensor product into a series of
smaller kernels with operands that fit entirely
into registers, enabling us to emit 
long arithmetic instruction streams that maximize 
instruction-level parallelism. By fusing the
CG tensor product with a subsequent 
graph convolution , we reduce both intermediate storage 
and global memory 
traffic over na\"{\i}ve approaches that duplicate
input data. We also provide optimized kernels
for the gradient of the CG tensor product and
a novel identity for the higher partial derivatives
required to predict interatomic forces. Our
kernels offer up to 1.3x speedup for the
over NVIDIA's closed-source cuEquivariance package, as
well as $>10$x speedup over the widely-used e3nn
package. In FP64 precision, we offer 
up to 6.2x inference-time speedup 
for the MACE chemistry 
foundation model over the original unoptimized version.

\end{abstract}

\section{Introduction}
Equivariant deep neural network models have 
become immensely popular in computational 
chemistry over the past seven years 
\cite{thomas_tensor_2018, steerable_cnns, cg_nets}. 
Consider a function 
$\bm f: \RR^n \rightarrow \RR^m$. Informally, $\bm f$ 
is \textit{invariant} if a class of transformations applied 
to its argument results in no change to the 
function output. A function is \textit{equivariant} if 
a transformation applied to any input argument of $\bm f$ can 
be replaced by a compatible transformation on the output 
of $\bm f$. For example: a function predicting molecular energy based on atomic positions should not change its result if the atom coordinates are rotated, translated, or 
reflected (invariance). Likewise, a function predicting 
3D forces on point masses should rotate its predictions 
if the input coordinate system rotates (equivariance). The latter property is termed \textit{rotation equivariance}, and it is the focus
of our work. Rotation equivariant 
neural architectures appear in 
the AlphaFold \cite{alphafold} version 2 model 
for protein structure prediction, the 
DiffDock \cite{corso2023diffdock} generative model 
for molecular docking, and
the Gordon Bell finalist Allegro \cite{allegro} for
supercomputer-scale molecular dynamics simulation,
among a host of other examples \cite{painnet, cormorant, batzner_2022,batatia2024foundationmodel, koker_chargenet}.
\begin{figure}
    \centering
    \includegraphics[width=1.0\linewidth]{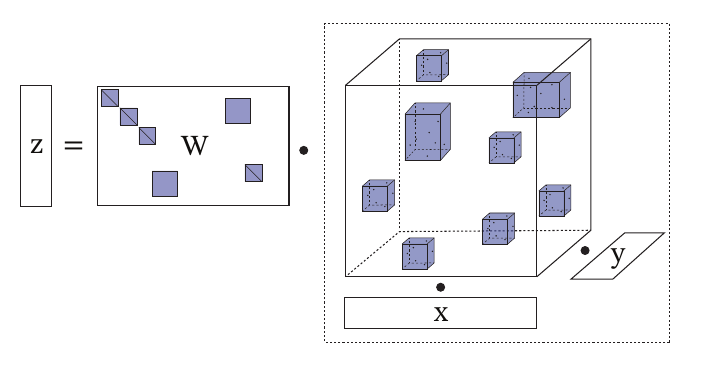}
    \caption{The CG tensor product, which 
    contracts a block-sparse tensor 
    $\scr P$ with two dense vectors to output a new vector. It is usually followed by 
    multiplication with a structured weight matrix $\bm W$, and by convention we use ``CG tensor product"
    to refer to both operations in sequence. Each
    blue block is itself sparse (see Figure
    \ref{fig:cg_tensor_structure}), and several blocks
    may share identical structure.}
    \label{fig:cg_tensor_product}
\end{figure}

A core kernel in many (though not all) rotation equivariant neural networks is the
Clebsch-Gordon (CG) tensor product, which combines two
feature vectors in an equivariant model 
to produce a new 
vector \cite{thomas_tensor_2018}. This multilinear operation, illustrated in
Figure \ref{fig:cg_tensor_product},
contracts a highly-structured block sparse tensor 
with a pair of dense input vectors, typically
followed by multiplication with 
a structured weight matrix. It is
frequently used to combine node and edge embeddings in
equivariant graph neural networks, which are used
for molecular energy prediction in computational chemistry 
\cite{batzner_2022, Batatia2022mace, allegro}
(see Figure \ref{fig:cg_tp_applications}).
With its low arithmetic intensity and irregular 
computation pattern, the CG tensor
product is difficult to implement efficiently in frameworks
like PyTorch or JAX. Because the CG tensor product and its 
derivatives must be evaluated millions of times 
on large atomic datasets, 
they remain significant bottlenecks to scaling equivariant neural networks. 

We introduce an open source kernel generator 
for the Clebsch-Gordon tensor product 
on both NVIDIA and AMD GPUs. Compared to the popular e3nn 
package \cite{e3nn_software} that is widely used in
equivariant deep learning models, 
we offer up to one order of magnitude improvement for both
forward and backward passes. 
Our kernels 
also exhibit up to 1.0-1.3x speedup 
over NVIDIA's closed-source cuEquivariance
v0.4.0 package \cite{geiger_accelerate_2024} on configurations used 
in graph neural networks, although our second derivative
kernel is slower by 30\% on certain inputs. 
Our key innovations include:
\newline

\paragraph{Exploiting ILP and Sparsity:}
 Each nonzero block of $\scr P$ in Figure \ref{fig:cg_tensor_product} is a structured sparse tensor (see
 Figure \ref{fig:cg_tensor_structure} for illustrations 
 of the nonzero pattern).
Popular existing codes \cite{e3nn_software}
fill these blocks with explicit zeros and use 
optimized dense linear algebra 
primitives to execute the tensor product,
performing unnecessary work in the process. By contrast,
we use Just-in-Time (JIT) compilation to generate
kernels that only perform work for nonzero tensor entries,
achieving significantly higher throughput as block sparsity 
increases. 
While previous works 
\cite{kondor_gelib, Koker_e3nn_c_2024}
use similar fine-grained approaches to optimize kernels
for each block in isolation, we achieve high 
throughput by JIT-compiling a single kernel for 
the entire sparse tensor $\scr P$. By compiling
kernels optimized for an entire sequence of nonzero blocks, we exhibit a degree of 
instruction level parallelism and data reuse
that prior approaches do not provide. 
\newline

\paragraph{Static Analysis and Warp Parallelism:} The structure of the sparse
tensor in Figure \ref{fig:cg_tensor_product}
is known completely at model 
compile-time and
contains repeated blocks with identical nonzero 
patterns. We perform a static analysis on the
block structure immediately after the
equivariant model architecture is defined to 
generate a computation schedule that 
minimizes global memory
traffic. We break the calculation into a series
of subkernels (see Figure \ref{fig:cg-primitives}),
each implemented by aggressively caching 
$\bm x$, $\bm y$, and $\bm z$ in the GPU register file.

We adopt a kernel design where each GPU warp operates 
on distinct pieces of coalesced data and requires
no block-level synchronization 
(e.g. \texttt{\_\_syncthreads()}). To accomplish this, each warp manages a unique 
portion of the shared memory pool and
uses warp-level matrix 
multiplication primitives to multiply
operands against nonzero blocks of the
structured weight matrix.
\newline

\paragraph{Fused Graph Convolution: } We demonstrate benefits far beyond reduced kernel launch overhead 
by fusing the CG tensor product with a 
graph convolution kernel (a very common pattern 
\cite{thomas_tensor_2018, batzner_2022, Batatia2022mace}).
We embed 
the CG tensor product and its backward pass 
into two algorithms 
for sparse-dense matrix multiplication: a simple, flexible
implementation using atomic operations and a faster
deterministic version using a fixup buffer. As a 
consequence, our work is the first to reap significant
model memory savings, a reduction in global memory writes,
and data reuse at the L2 cache level. 
\newline

Section
\ref{sec:preliminaries} provides a brief introduction 
to equivariant neural networks, with Section 
\ref{sec:core_challenge} providing a precise definition
and motivation for the CG tensor product. Section
\ref{sec:engineering_cg_kernels} details our strategy
to generate efficient CG kernels and the design decisions
that yield high performance. We validate those decisions
in Section \ref{sec:experiments} on a range of benchmarks
from chemical foundation models and other 
equivariant architectures.

\section{Preliminaries and Problem Description}\label{sec:preliminaries}

We denote vectors, matrices, and
three-dimensional tensors in
bold lowercase characters, bold uppercase
characters, and script characters (e.g.
$\scr P$), respectively. Our notation and description of equivariance 
follow \citet{thomas_tensor_2018} and
\citet{lim_what_2023}. Let 
$G$ be an abstract group of transformations, and let
$\bm D_{\textrm{in}}: G \rightarrow \RR^{n \times n}, \bm D_{\textrm{out}}: G
\rightarrow \RR^{m \times m}$ be a pair 
of \textit{representations}, group homomorphisms satisfying 
\[
\bm D_{\textrm{in}}(g_1 \cdot g_2) = \bm D_{\textrm{in}}(g_1) \cdot \bm D_{\textrm{in}}(g_2)\quad \forall g_1, g_2 \in G,
\]
and likewise for $\bm D_{\textrm{out}}$. A function $\bm f: \RR^n \rightarrow \RR^{m}$ is equivariant
with respect to $\bm D_{\textrm{in}}$ and $\bm D_{\textrm{out}}$ iff 
\[\bm f(\bm D_{\textrm{in}}(g) \cdot \bm v) = \bm D_{\textrm{out}}(g) \cdot \bm f(\bm v)\quad\forall \bm v \in \RR^n, g \in G.
\]
A function is invariant if the equivariance property holds with
$\bm D_{\textrm{out}}(g) = \bm I^{m \times m}$ for all $g \in G$. 

In our case, $\bm f$ is a neural network composed 
of a sequence of layers, expressed
as the function composition
\[
\bm f(\bm v) = \bm \phi_{N} \circ ... \circ \bm \phi_1(\bm v).
\]
Here, $\bm D_{\textrm{in}}$ and $\bm D_{\textrm{out}}$ are
derived from the dataset, and the task is to
fit $\bm f$ to a set of datapoints while 
maintaining equivariance to the 
chosen representations. Network designers 
accomplish this by imposing equivariance on
each layer and exploiting a composition
property \cite{thomas_tensor_2018}: if $\bm \phi_i$ is
equivariant to input / output representations 
$(\bm D_i, \bm D_{i+1})$ and
$\bm \phi_{i+1}$ is equivariant to $(\bm D_{i+1}, \bm D_{i+2})$, then
$\bm \phi_{i+1} \circ \bm \phi_i$ is equivariant to 
$(\bm D_i, \bm D_{i+2})$. These intermediate representations
are selected by the network designer to maximize predictive capability. 

\subsection{Representations of O(3)}
\label{sec:o3_reps}

\begin{figure}
    \centering
    \includegraphics[width=1\linewidth]{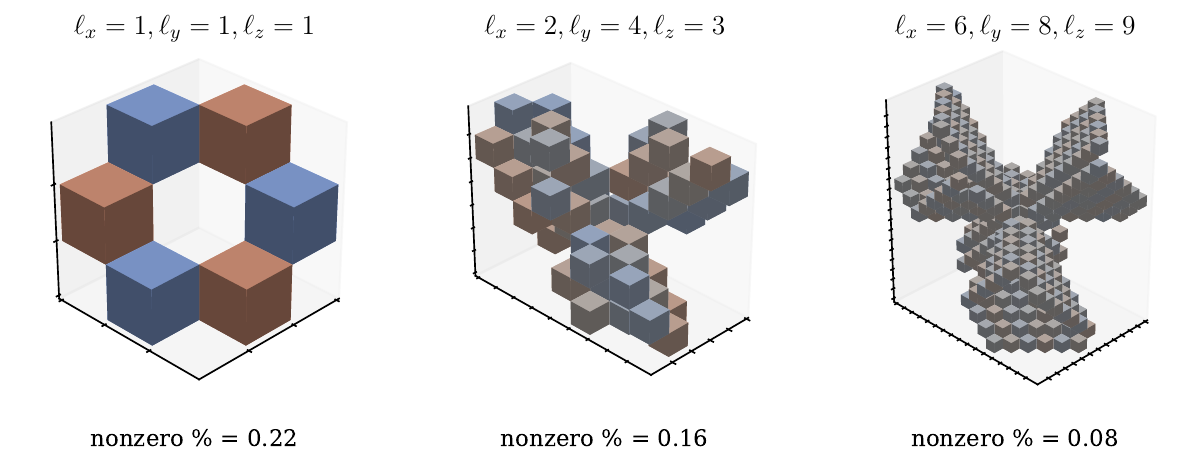}
    \caption{Three examples of coefficient 
    tensors depicted by blue cubes in Figure
    \ref{fig:cg_tensor_product}. The fraction of zero 
    entries increases with the tensor order, and the full CG tensor contains several 
    copies of the blocks pictured here.}
    \label{fig:cg_tensor_structure}
\end{figure}

In this paper, we let $G = O(3)$, the group of 
three-dimensional rotations including reflection. 
A key property of real representations
of $O(3)$ is our ability to block-diagonalize them into a 
canonical form \cite{luo2024gaunt}. Formally, for any representation
$\bm D: O(3) \rightarrow \RR^{n \times n}$ and all
$g \in G$, there exists a
similarity matrix $\bm P$ and indices $i_1, ..., i_D$ satisfying 
\begin{equation*}
\begin{aligned}    
\bm D(g) = \bm P^{-1}
\begin{bmatrix}
\bm D^{(i_1)}(g) & & 0\\
& \ddots & \\
0& & \bm D^{(i_D)}(g)
\end{bmatrix} \bm P
\end{aligned}
\end{equation*}
where $\bm D^{(0)}(g), \bm D^{(1)}(g), ...$ are a family of
elementary, \textit{irreducible} representations known as the Wigner
D-matrices. For all $i \geq 0$, we have
$\bm D^{(i)}(g) \in \RR^{(2i+1) \times (2i+1)}$. In 
the models we consider, all
representations will be exactly block diagonal
(i.e. $\bm P$ is the identity matrix), described
by strings
of the form
\[
\bm D(g) \cong \texttt{"3x1e + 1x2o"}.
\]
This notation indicates that $D$ has 
three copies of $D^{(1)}$ along the diagonal
followed by one copy of $D^{(2)}$. We refer to the term 
\texttt{"3x1e"} as an irrep (irreducible representation) 
with $\ell=1$ and multiplicity 3. 
The suffix letters, ``e"
or ``o", denote a parity used to enforce
reflection equivariance, which is not relevant for us 
(we refer the reader to \citet{thomas_tensor_2018} for
a more complete explanation).

\subsection{Core Computational Challenge}
\label{sec:core_challenge}
Let $\bm x \in \RR^n, \bm y \in \RR^m$ 
be two vectors from some 
intermediate layer $\bm \phi$ of an equivariant
deep neural network. For example, $\bm x$ 
could be the embedding associated 
with a node of a graph and $\bm y$ a feature vector
for an edge (see Figure \ref{fig:cg_tp_applications}, bottom). 
We can view both vectors 
as functions $\bm x(\bm v), \bm y(\bm v)$ 
of the network input $\bm v$, which are
equivariant to 
$(\bm D_{\textrm{in}}, \bm D_x)$ and 
$(\bm D_{\textrm{in}}, \bm D_y)$ respectively. 
An equivariant graph convolution layer $\bm \phi$ 
interacts $\bm x$ and $\bm y$ to produce
a new vector $\bm z$. To ensure
layer-equivariance of $\phi$, $\bm z(\bm v)$ must be equivariant
to $(\bm D_{\textrm{in}}, \bm D_z)$, where $\bm D_z$ is
a new representation selected by the network
designer. 

The Kronecker product provides an expressive, general method to interact $\bm x$ and $\bm y$: if $\bm x(\bm v)$ and $\bm y(\bm v)$
are equivariant to the representations listed above,
then $\bm z(\bm v) = \bm x(\bm v) \otimes \bm y(\bm v)$ 
is equivariant to
$(\bm D_{\textrm{in}},\bm D_x \otimes \bm D_y)$. 
Unfortunately, 
$\bm x \otimes \bm y \in \RR^{nm}$ may have intractable length, and we
cannot drop arbitrary elements
of the Kronecker product without
compromising the equivariance property.

Let $\bm P \in \RR^{nm \times nm}$ be the similarity transform 
diagonalizing $\bm D_x \otimes \bm D_y$.
To reduce the dimension of the Kronecker
product, we first form 
$\bm P (\bm x(\bm v) \otimes \bm y(\bm v))$, an equivariant 
function with block-diagonal output
representation $\bm P(\bm D_x \otimes \bm D_y)\bm P^{-1}$. We can now safely remove segments of 
$\bm P (\bm x \otimes \bm y)$ corresponding to unneeded higher-order
Wigner blocks and recombine its 
components through a trainable, structured
weight matrix. The result, $\bm z(\bm v) = \bm W \bm P (\bm x(\bm v) \otimes \bm y(\bm v))$, 
has a new output representation 
$\bm D_z$ and can be much shorter than $\bm x \otimes \bm y$. 

When both $\bm D_x$ and $\bm D_y$ are representations in block-diagonal canonical form, the transform $\bm P$ is
a highly structured block-sparse matrix containing 
nonzero \textit{Clebsch-Gordon coefficients}. 
After potentially reducing $\bm P$ to $k$ rows 
(removing segments corresponding
to unneeded Wigner D-blocks), we can
reshape it into a block-sparse tensor $\scr P \in 
\RR^{m \times n \times k}$ contracted on two sides with
$\bm x$ and $\bm y$. We call this operation (along with multiplication
by a structured weight matrix $\bm W \in \RR^{k \times k'}$) 
the \textbf{CG tensor product}, illustrated in 
Figure \ref{fig:cg_tensor_product}. It can be expressed 
by a matrix equation, a summation expression, 
multilinear tensor contraction
(popular in the numerical linear algebra community), 
or Einstein notation:
\begin{equation}
\begin{aligned}    
\bm z
&= \boxed{\tp(\scr P, \bm x, \bm y, \bm W)} \\
&:= \bm W \cdot \bm P \cdot (\bm x \otimes \bm y) \\
&:= \bm W \sum_{i=1,j=1}^{m, n} \bm x\br{i} \bm y\br{j} \scr P\br{ij:} \\ 
&:= \scr P \times_1 \bm x \times_2 \bm y \times_3 \bm W \\
&:= \textrm{einsum}(``ijk,i,j,kk'\rightarrow k'", \scr P, \bm x, \bm y, \bm W).
\end{aligned}
\label{eq:cg_definition}
\end{equation}
Our goal is to accelerate computation of 
$\tp(\scr P, \bm x, \bm y, \bm W)$ for a variety of CG coefficient tensors
$\scr P$. Given $\partial E / \partial \bm z$ for some scalar
quantity $E$, we will also provide an efficient kernel to
compute $\partial E / \partial \bm x$, $\partial E / \partial \bm y$,
and $\partial E / \partial \bm W$ in a single pass. These 
gradients are required during both training and inference for some interatomic potential models. 

\subsection{Structure in the Sparse Tensor and
Weights}

\begin{figure}
    \centering    \includegraphics[width=1.00\linewidth]{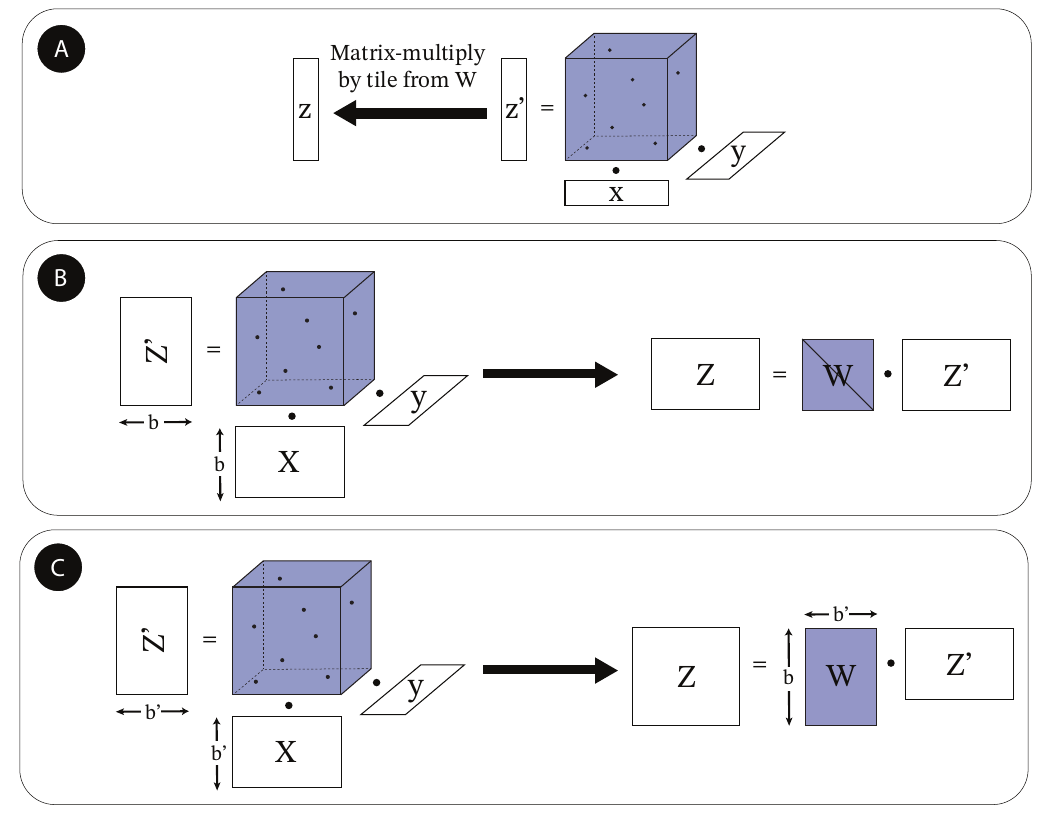}
    \caption{Fundamental subkernels that
    compose to implement the CG tensor
    product in Figure \ref{fig:cg_tensor_product}. In (A), 
    $\bm x$, $\bm y$, and $\bm z$ refer to segments
    of the longer vectors in Figure \ref{fig:cg_tensor_product}, and $\bm W$
    contains entries from the larger
    weight matrix rearranged 
    appropriately. In (B) and
    (C), $\bm X$ and $\bm Z$ refer to segments from
    $\bm x$ and $\bm z$ that have been reshaped
    into matrices to exploit repeating sparse tensor structure.
    $b$ and $b'$ are multiples of 32 in many models.}
    \label{fig:cg-primitives}
\end{figure}

Suppose $\bm D_x$, 
$\bm D_y$, and $\bm D_z$ each consist of a 
single Wigner block.
In this case, the tensor $\scr P$
in Figure \ref{fig:cg_tensor_product}
has a single nonzero block
of dimensions
$(2\ell_x + 1) \times (2 \ell_y + 1) \times (2\ell_z + 1)$.
Figure 
\ref{fig:cg_tensor_structure} 
illustrates three blocks with varying
parameters, which are small,
(current models typically use 
$\ell_x, \ell_y, \ell_z \leq 4$), highly structured, and sparse. 

Every nonzero 
block of the general tensor $\scr P$ in
Figure \ref{fig:cg_tensor_product} takes
the form $\scr P^{(\ell_x,
\ell_y, \ell_z)}$. Therefore, we could implement the
CG tensor product by repeatedly calling 
the kernel in Figure \ref{fig:cg-primitives}A: a small
tensor contraction followed by multiplication
by a tile from the weight matrix $\bm W$. In practice, this is an inefficient strategy because $\scr P$ may contain hundreds of blocks 
with identical nonzero
structures and values. 

Instead, the CG tensor product splits 
into a sequence of subkernels \cite{kondor_gelib} that
match the structure in $\scr P$ and $\bm W$. We target two
common patterns. First, the 
CG tensor product 
may interact $b$ unique segments of $\bm x$ with a common
segment of $\bm y$ using the same 
block from $\scr P$,
followed by multiplication by a submatrix of weights from
$\bm W$ rearranged along a diagonal. 
Figure \ref{fig:cg-primitives}B illustrates 
Kernel B as a contraction of a sparse tensor with
a matrix $\bm X$ (containing the $b$ rearranged segments from $\bm x$)
and the common vector $\bm y$. It appears
in the Nequip \cite{batzner_2022}
and MACE \cite{Batatia2022mace} 
models. 
Kernel C is identical to kernel B, but
arranges the weights in a dense matrix 
$\bm W \in \RR^{b' \times b}$. Here, $\bm Z$ and $\bm X$ may have
distinct row counts. The latter operation appears in DiffDock
\cite{corso2023diffdock} and 3D 
shape classifiers by \citet{thomas_tensor_2018}.
Both operations can be extended to interact
multiple segments from $\bm y$, but we
could not find this pattern in
existing equivariant models. While
kernels besides B and C are possible,
they rarely appear in practice. 

\subsection{A Full Problem Description}
\label{sec:example_problem}
Armed with the prior exposition, we now give 
an example of a specific CG tensor product using the
notation of the e3nn software package
\cite{e3nn_software}. A specification 
of the tensor product consists of a sequence
of subkernels and the representations that 
partition $\bm x$, $\bm y$, and $\bm z$ into segments for those kernels to operate on: 
\begin{equation}
\begin{aligned}
&\bm D_x \cong \texttt{32x2e + 32x1e} \\
&\bm D_y \cong \texttt{1x3e + 1x1e} \\
&\bm D_z \cong \texttt{32x5e + 16x2e + 32x3e} \\
&\br{(1, 1, 1, ``B"),
(1, 2, 2, ``C"), (1, 2, 3, ``C")
}.
\end{aligned}
\label{eq:example_problem}
\end{equation}
Here, $\bm x$ and $\bm y$ are partitioned into
two segments, while $\bm z$ is partitioned into three
segments. The list of tuples on the last line 
specifies the subkernels to execute
and the segments they operate on. The first list 
entry specifies kernel B with the 
respective first segments  
of $\bm x$, $\bm y$, and $\bm z$ ($b = 32$). Likewise, the second
instruction executes kernel C
with the first segment of $\bm x$ and 
the second segment of $\bm y$ ($b'=32,b=16$) to produce the second 
segment of $\bm z$. 

\subsection{Context and Related Work}
Variants of rotation equivariant 
convolutional neural
networks were first proposed 
by \citet{steerable_cnns};
\citet{cg_nets}; and \citet{thomas_tensor_2018}. 
Nequip \cite{batzner_2022}, Cormorant \cite{cormorant}, and 
Allegro \cite{allegro} deploy equivariant
graph neural networks to achieve state-of-the-art
performance for molecular energy prediction.
Other works have enhanced these message-passing architectures 
by adding higher-order equivariant features (e.g.
MACE \cite{Batatia2022mace, batatia2024foundationmodel} or ChargE3Net \cite{koker_chargenet}). Equivariance has been
integrated into transformer architectures
\cite{fuchs_transformer, equiformer} with similar
success.
\newline

\paragraph{Equivariant Deep Learning Software} The e3nn package
\cite{e3nn_paper, e3nn_software} allows
users to construct CG interaction tensors, compute CG tensor products,
explicitly form Wigner D-matrices, and 
evaluate spherical harmonic basis functions. PyTorch and JAX versions of the package
are available. The e3x package 
\cite{e3x_equivariant_dl} offers
similar functionality. Allegro \cite{allegro}
modifies the message passing equivariant
architecture in Nequip \cite{batzner_2022}
to drastically reduce the number of CG tensor 
products and lower inter-GPU communication. 

The cuEquivariance package \cite{geiger_accelerate_2024}, released by NVIDIA 
concurrent to this project's development, offers 
the fastest implementation of the CG tensor product
outside of our work. However, their
kernels are closed-source and do not appear to exploit
sparsity \textit{within} each nonzero block of $\scr P$. Our kernel performance
matches or exceeds cuEquivariance, often by substantial margins. Other relevant
codes include GELib 
\cite{kondor_gelib}, Sphericart 
\cite{sphericart}, and Equitriton \cite{lee2024scaling}. The former
efficiently computes CG tensor products, while
the latter two accelerate 
spherical harmonic polynomial evaluation.
\newline

\paragraph{Alternatives to CG Contraction}
The intense cost of the CG tensor product has
fueled the search for cheap yet accurate 
algorithms. For example, \citet{painnet} propose an
equivariant message passing neural network
that operates in Cartesian space, eliminating
the need for CG tensor products. For 
$\textrm{SO(3)}$-equivariant graph 
convolution, \citet{pz_so3_so2} 
align the node embeddings with spherical harmonic edge features before interacting the two. This
innovation sparsifies the interaction tensor and asymptotically
decreases the cost of each tensor product. \citet{luo2024gaunt} also produce asymptotic speedups by connecting the tensor product 
with a spherical harmonic product accelerated through 
the fast Fourier transform, an operation they
call the \textit{Gaunt tensor product}.
\citet{xie_price_2024} counter that the Gaunt
tensor product does not produce results directly
comparable to the CG tensor product, and
that the former may sacrifice model expressivity. 
\newline

\paragraph{GPU Architecture} 
GPUs execute kernels by launching 
a large number of parallel threads running the same
program, each accessing a small set of local registers. 
In a typical program, threads load data from global
memory, perform computation with data in their registers,
and store back results. Kernels
are most efficient when groups of 32-64 threads (called ``warps") execute the same instruction or perform memory
transaction on contiguous, aligned segments of data. 
Warps execute asynchronously and are grouped 
into cooperative thread arrays (CTAs) that communicate through a fast, limited pool of shared memory. Warps can 
synchronize at the CTA level, but 
the synchronization incurs overhead.

\section{Engineering Efficient CG Kernels}
\label{sec:engineering_cg_kernels}
\begin{figure}
    \centering
    \includegraphics[width=1.0\linewidth]{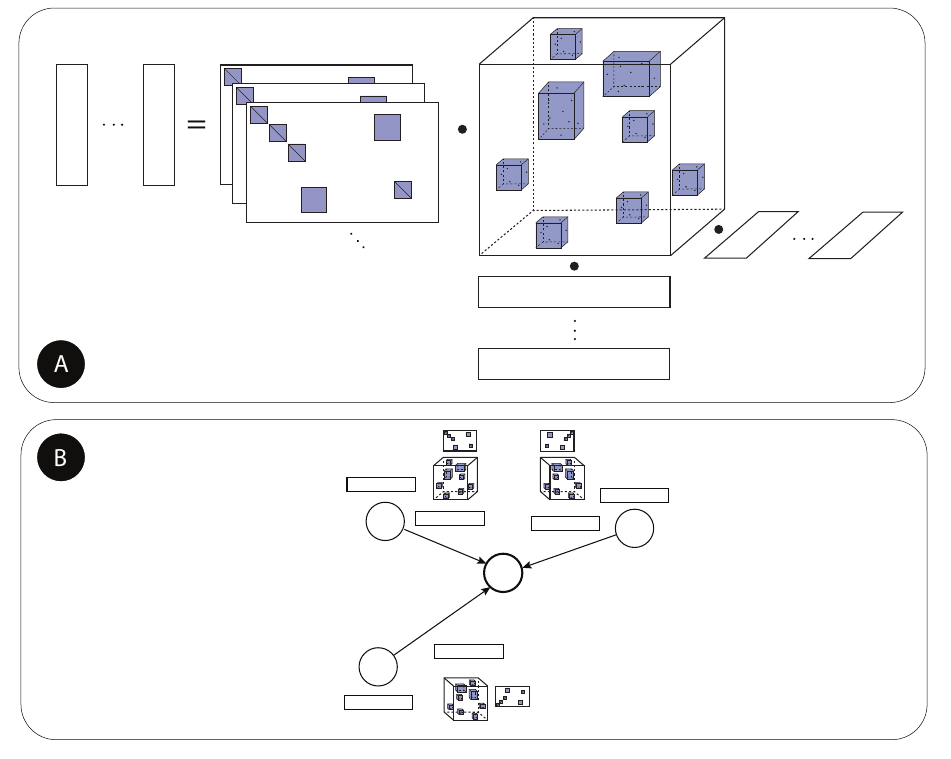}
    \caption{Applications of the CG tensor product. The simplest and most general use case 
    (A) calls the kernel repeatedly with distinct 
    $\bm x$, $\bm y$, and $\bm W$ inputs. Interatomic potential models
    embed the operation in
    a graph convolution (B), where the tensor product
    combines node features with edge features.}
    \label{fig:cg_tp_applications}
\end{figure}

Our task is to generate an efficient CG
tensor product kernel given a problem specification 
outlined in Section \ref{sec:example_problem}. Algorithm \ref{alg:cg_template} describes the logic 
of a kernel that operates on a large batch of inputs, each with a distinct set of
weights (see Figure \ref{fig:cg_tp_applications}A). We 
assign each $(\bm x, \bm y, \bm W)$ input triple to a single GPU warp, a choice which has two consequences.
First, it enables each warp to execute contiguous global memory 
reads / writes for $\bm x, \bm y, \bm W$ and $\bm z$. Second, it allows 
warps to execute in a completely asynchronous fashion without
any CTA-level synchronization, boosting throughput 
significantly. The weights $\bm W$ are
stored in a compressed form without
the zero entries illustrated in
Figure \ref{fig:cg_tensor_product}.

After the division of labor, each warp follows
a standard GPU kernel workflow. The three
inputs are staged in shared memory, 
the kernels in Equation 
\eqref{eq:example_problem} are executed sequentially, and each output $\bm z_b$ is stored back. Each warp operates on a unique 
partition of the CTA shared memory which may not be large
enough to contain the the inputs and outputs. In 
the latter 
case, chunks of $\bm x$, $\bm y$, $\bm W$, and $\bm z$ 
are staged, and the computation executes in phases according to a schedule
described in 
Section \ref{sec:computation_schedule}.

\begin{algorithm}
\caption{High-Level CGTP Algorithm}
\begin{algorithmic}
\Require Batch $\bm x_1...\bm x_B$, $\bm y_1...\bm y_B$,$\bm W_1...\bm W_B$

\For{$b=1...B$} \Comment{Parallel over warps}
    \For{$\textrm{seg}_i \in \textrm{schedule}$}
        \State Load $\bm x_{\textrm{smem}} = 
        \bm x_b\br{\textrm{seg}_{i\ \textrm{(x start)}}:
        \textrm{seg}_{i\ \textrm{(x end)}}}$
        \State Load $\bm y_{\textrm{smem}}, \bm W_{\textrm{smem}}$ similarly.
        \State Set $\bm z_{\textrm{smem}} = 0$
        \For{$\textrm{kern}_j \in \textrm{seg}_i$}
            \State Set $\bm X_{\textrm{kern}}$ as
            a reshaped range of $\bm x_{\textrm{smem}}$
            \State Set $\bm y_{\textrm{kern}}, \bm W_{\textrm{kern}}$ similarly.
            \State 
            $\bm Z_\textrm{kern} = \textrm{kern}_j(\bm X_{\textrm{kern}}, \bm y_{\textrm{kern}}, \bm W_{\textrm{kern}})$
            \State Flatten, store 
            $\bm Z_{\textrm{kern}}$ to subrange of
            $\bm z_{\textrm{smem}}$
        \EndFor 
        \State Store $
        \bm z_b\br{\textrm{seg}_{i\ \textrm{(z start)}}:
        \textrm{seg}_{i\ \textrm{(z end)}}} \pluseq 
        \bm z_{\textrm{smem}}$
    \EndFor
\EndFor
\end{algorithmic}
\label{alg:cg_template}
\end{algorithm}
We launch a number of CTAs that is a constant multiple of the
GPU streaming multiprocessor count 
and assign 4-8 warps per
CTA. The batch size for the
CG tensor product can reach millions \cite{allegro} 
for large geometric configurations, ensuring that all warps 
are busy. The computation required for each batch element 
can exceed 100K FLOPs for typical models 
\cite{allegro, batzner_2022}, ensuring that 
the threads within each warp are saturated with work. 

\subsection{Computation Scheduling}
\label{sec:computation_schedule}
A key obstacle to efficient kernel
implementation is the long length of the $\bm x$, $\bm y$, and $\bm z$  
feature vectors that must be cached in shared memory. The sum of their vector lengths for large
MACE \cite{batatia2024foundationmodel} 
and Nequip \cite{batzner_2022} configurations can 
exceed 10,000 data words.
Given that the warps in each CTA partition the shared memory,
staging all three vectors at once (along with the 
weights in $\bm W$) is infeasible.

To manage our limited shared memory, we
execute the computation in phases that
are scheduled at model compile-time. We break the list of
instructions in Equation \eqref{eq:example_problem} into
phases so that the sum of chunks from $\bm x$, $\bm y$, $\bm W$ and 
$\bm z$ required for the phase fits 
in each warp's shared memory allotment. We then schedule
loads and stores, hardcoding the relevant instructions 
into each kernel using our JIT capability. When more than a single computation
phase is required, our goal is to generate a schedule that minimizes global 
memory reads and writes. We use a few
simple heuristics:
\begin{enumerate}
\item If $\bm x$ and $\bm y$ can fit into a warp's
shared memory partition 
(but not $\bm z$ and
$\bm W$), then
segments of $\bm z$ and $\bm W$ are streamed in through
multiple phases of computation. In each phase, 
the kernels that touch each segment of $\bm z$ are
executed. 

\item Otherwise, we use a greedy
algorithm. In each phase, the
shared memory pool is filled with as many segments
of $\bm x$, $\bm y$, $\bm W$ and $\bm z$ that can fit. 
Segments are 
flushed and reloaded as needed.
\end{enumerate}

Case 1 covers 
most problem configurations in
equivariant graph neural networks and minimizes global memory
writes, while Case 2
enables reasonable performance even with 
constrained shared memory resources. Large CG tensors 
(e.g. Nequip-benzene in Figure \ref{fig:uvu_throughput_comparison}) may require 20-40
phases of computation per tensor product, 
and our scheduling substantially 
reduces global memory transactions.

\subsection{JIT Subkernel Implementations }
\label{sec:kernel_implementations}
In preparation to execute a subkernel,
suppose we have 
loaded $\bm x$, $\bm y$ and $\bm W$ into shared
memory and reshaped subranges of
all three to form $\bm X_{\textrm{kern}},
\bm y_{\textrm{kern}}$, 
and $\bm W_{\textrm{kern}}$. For the rest
of Section \ref{sec:kernel_implementations}
and all of Section \ref{sec:backward_pass}, we omit the ``kern" subscripts. Algorithm 
\ref{alg:kernel_bc} gives the
pseudocode to execute either kernel B or C
from Figure \ref{fig:cg-primitives}
using these staged operands.
\begin{algorithm}
\caption{Subkernel B \& C Warp-Level Algorithm}\label{alg:kernel_bc}
\begin{algorithmic}
\Require $\bm X \in \RR^{b' \times (2\ell_x+1)}, 
\bm y \in \RR^{(2 \ell_y + 1)}, \bm W \in \RR^{b \times b'}$ 

\Require Sparse tensor $\scr P^{(\ell_x, \ell_y, \ell_z)}$ for subkernel

\For{$t=1...b'$} \Comment{Parallel over threads}
    \State Load $\bm x_{\textrm{reg}} = \bm X\br{t, :}$,
    $\bm y_{\textrm{reg}} = \bm y$ 
    
    \State Initialize 
    $\bm z_{\textrm{reg}} \in \RR^{2\ell_z + 1}$ to 0.

    \State
    
    \For{$(i, j, k, v) \in \textrm{nz}(\scr P )$} \Comment{Unroll via JIT} 
    \State $\bm z_\textrm{reg}\br{k} \pluseq v \cdot \bm x_\textrm{reg}\br{i} \cdot \bm y_\textrm{reg}\br{j}$
    \EndFor
    \State
    
    \If{$\bm W$ is diagonal} \Comment{Compile-time branch}
        \State $\bm Z\br{t, :} \pluseq \bm W\br{t, t} \cdot \bm z_\textrm{reg}$
    \Else
        \State Store $\bm Z'\br{t, :} = \bm z_{\textrm{reg}}$, compute $\bm Z \pluseq \bm W \cdot \bm Z'$
    \EndIf
\EndFor
\end{algorithmic}
\end{algorithm}
Each thread stages a unique row
row of $\bm X$ and $\bm Z$, as well as the entirety 
of $\bm y$, into its local registers. 
Models such as Nequip 
\cite{batzner_2022} and MACE 
\cite{batatia2024foundationmodel} 
satisfy $\ell_x, \ell_y, \ell_z \leq 4$, so the 
added register pressure from the operand
caching is manageable. We then loop over 
all nonzero entries of the sparse tensor to execute the
tensor contraction. Because the nonzero indices $(i, j, k)$ and entries $\bm v$
of the sparse tensor $\scr P$ 
are known at compile-time,
we emit the sequence of instructions in the
inner loop explicitly using our JIT kernel 
generator. Finally, the output $\bm Z$ is 
accumulated to shared memory after 
multiplication by $\bm W$.
When multiple 
subkernels execute in sequence,
we allow values in $\bm x_\textrm{reg}$, $\bm y_\textrm{reg}$, and $\bm z_\textrm{reg}$ to persist if they are reused.

The matrix multiplication by the weights 
at the end of Algorithm \ref{alg:kernel_bc} 
depends on the structure 
of $\bm W$. When $\bm W$ is square and diagonal (kernel B),
multiplication proceeds asynchronously
in parallel across all threads. When $\bm W$
is a general dense matrix, we temporarily store
$\bm z_{\textrm{reg}}$ to shared memory and
perform a warp-level matrix-multiplication across
all threads.

Our kernel generator maximizes instruction-level
parallelism, and the output kernels 
contain long streams of independent 
arithmetic operations. By contrast,
common sparse tensor storage formats 
(coordinate \cite{alto_storage_format}, compressed-sparse fiber \cite{smith_csf}, etc.) require expensive
memory indirections that reduce throughput. 
Because we compile a single kernel to handle 
all nonzero blocks of $\scr P$, we 
avoid expensive runtime branches and permit data
reuse at the shared memory and register level. Such
optimizations would be difficult to implement in
a traditional statically-compiled library.

For typical applications, $b$
and $b'$ are 
multiples of 32. When $b$ is greater than 32,
the static analysis algorithm in Section
\ref{sec:computation_schedule} breaks the computation
into multiple subkernels with $b \leq 32$, and likewise for $b'$.

\subsection{Backward Pass}
\label{sec:backward_pass}
Like other kernels in physics informed deep learning
models \cite{pinn_nature_review}, the
gradients of the CG tensor product are 
required during model \textit{inference} as 
well as training
for interatomic force prediction. Suppose 
$E(\bm R, \bm W)$ is the scalar
energy prediction emitted by our 
equivariant model for a 
configuration of $s$ atoms,
where $\bm W$ contains trainable model
weights and each row of $\bm R \in \RR^{s \times 3}$
is an atom coordinate. Then 
$\bm F_{\textrm{pr}} = - \partial E / \partial \bm R \in \RR^{s \times 3}$ is the 
predicted force on each atom. Conveniently, we can compute 
these forces by auto-differentiating 
$E(\bm R, \bm W)$ in a framework like PyTorch or JAX,
but we require a kernel to compute the gradient
of the CG tensor product inputs given the
gradient of its output.

To implement the backward pass, suppose
$\bm z = \tp(\scr P, \bm x, \bm y, \bm W)$ and we have 
$\bm g_z = \partial E / \partial \bm z$. Because the
CG tensor product is linear in its inputs, 
the product rule gives
\begin{equation*}
\begin{aligned}
\partial E / \partial \bm x\br{i} &=
\sum_{(i, j, k) \in \textrm{nz}(\scr P)}
\scr P\br{ijk} \cdot \bm y\br{j} \cdot 
\paren{\bm W^\top \cdot \bm g_z}\br{k}
 \\
\partial E / \partial \bm y\br{j} &=
\sum_{(i, j, k) \in \textrm{nz}(\scr P)}
\scr P\br{ijk} \cdot \bm x\br{i} \cdot 
\paren{\bm W^\top \cdot \bm g_z}\br{k}
 \\ 
\partial E / \partial \bm W\br{kk'} &= 
\bm g_z \br{k'} \cdot \sum_{(i,j,k) \in \textrm{nz}(\scr P)} \scr P \br{ijk} \bm x\br{i} \bm y\br{j} \\
\end{aligned}
\end{equation*}
Notice the similarity between the three 
equations above and 
Equation \eqref{eq:cg_definition}: 
all require
summation over the nonzero indices of $\scr P$
and multiplying each value with a pair of elements
from distinct vectors. Accordingly, we develop
Algorithm \ref{alg:kernel_b_backward} with 
similar structure to Algorithm \ref{alg:kernel_bc}
to compute all three gradients in a single kernel. For simplicity,
we list the general case where the submatrix 
$\bm W$ is a general dense matrix
(kernel C).
\begin{algorithm}
\caption{Subkernel C Warp-Level Backward}\label{alg:kernel_b_backward}
\begin{algorithmic}
\Require $\bm X \in \RR^{b' \times (2\ell_x + 1)}, 
\bm y \in \RR^{2\ell_y + 1}, \bm W \in \RR^{b \times b'}$
\Require $\bm G_Z \in \RR^{b \times (2\ell_z + 1)}$,
sparse tensor $\scr P^{(\ell_x, \ell_y, \ell_z)}$

\State Threads collaboratively compute $\bm G_Z' = \bm W^\top \cdot \bm G_Z$
\For{$t=1...b'$} \Comment{Parallel over threads}
    \State Load $\bm x_{\textrm{reg}} = \bm X\br{t, :}$,
    $\bm y_{\textrm{reg}} = \bm y$, $\bm g_{z \textrm{reg}}' = \bm G_Z'\br{t:}$

    \State Initialize $\bm g_{\textrm{x reg}}, \bm g_{\textrm{y reg}}, \bm g_{\textrm{w reg}}, \bm z_{\textrm{reg}}$ to 0.
    \State 
    \For{$(i, j, k, v) \in \textrm{nz}(\scr P^{(\ell_x, \ell_y, \ell_z)}$} \Comment{Unroll via JIT} 
    \State $\bm g_\textrm{x reg}\br{i} \pluseq 
    v \cdot \bm y_\textrm{reg}\br{j} \cdot \bm g_\textrm{z reg}'\br{k}$ 
    \State $\bm g_\textrm{y reg}\br{j} \pluseq 
    v \cdot \bm x_\textrm{reg}\br{i} \cdot \bm g_\textrm{z reg}'\br{k}$  
    \State $\bm z_\textrm{reg}\br{k} \pluseq v \cdot \bm x_\textrm{reg}\br{i} \cdot \bm y_\textrm{reg}\br{j}$
    \EndFor
    \State
    \State Store $\bm g_y = \textrm{warp-reduce}(\bm g_{\textrm{y reg}})$ 
    \State Store $\bm G_x\br{t, :} = \bm g_{\textrm{x reg}}$ and $\bm Z'\br{t, :} = \bm z_{\textrm{reg}}$
\EndFor
\State Threads collaboratively compute $G_W = \bm G_Z \cdot (\bm Z')^\top$
\end{algorithmic}
\end{algorithm}
There are two new key features in 
Algorithm \ref{alg:kernel_b_backward}:
first, we must perform a reduction over the warp
for the gradient vector $\bm g_y$, since each thread
calculates a contribution that must be summed. Second:
when $\bm W$ is not diagonal, an additional warp-level matrix multiply
is required at the end of the algorithm to calculate $\bm G_W$. We embed
Algorithm \ref{alg:kernel_b_backward} into a high-level procedure
akin to Algorithm \ref{alg:cg_template} to complete the
backward pass.

\subsection{Higher Partial Derivatives}
For interatomic potential 
models, we require higher-order derivatives 
to optimize force predictions during training 
\cite{pinn_nature_review}, as 
we explain below. Rather than 
write new kernels for these derivatives, 
we provide a novel 
(to the best of our knowledge) calculation
that implements them using the existing
forward and backward pass kernels.

As in Section \ref{sec:backward_pass}, let 
$\bm F_{\textrm{pr}} = - \partial E / \partial \bm R \in \RR^{s \times 3}$ be the predicted atomic 
forces generated by our model. During 
training, we must minimize a loss function of 
the form
\[
\min_{\bm W} \scr L(\bm R, \bm W) = \min_{\bm W} \norm{\bm F_{\textrm{pr}}(\bm R, \bm W) - \bm F_{\textrm{gt}}(\bm R)}_{F}^2
\]
where 
$\bm F_{\textrm{gt}}(\bm R) \in \RR^{s \times 3}$ is
a set of ground-truth forces created from a more 
expensive simulation. The loss function may include
other terms, but only the Frobenius norm of the 
force difference is relevant here. We use a gradient
method to perform the minimization and calculate 
\begin{equation}
\begin{aligned}    
\frac{\partial \scr L}{\partial \bm W} 
&=2 \cdot \textrm{vec}(\bm F_{\textrm{pr}}(\bm R, \bm W) - \bm F_{\textrm{gt}}(\bm R))^\top \frac{\partial \bm F_{\textrm{pr}}}{\partial \bm W} \\
&=-2 \cdot \textrm{vec}(\bm F_{\textrm{pr}}(\bm R, \bm W) - \bm F_{\textrm{gt}}(\bm R))^\top \frac{\partial^2 E}{\partial \bm R \partial \bm W}
\end{aligned}
\label{eq:second_partial}
\end{equation}
where ``$\textrm{vec}$" flattens its matrix argument
into a vector and $\partial^2 E / (\partial \bm R \partial \bm W) \in \RR^{3s \times (\textrm{\# weights})}$ is a matrix of second 
partial derivatives. Equation \eqref{eq:second_partial}
can also be computed by auto-differentiation,
but the second partial derivative requires us
to register an autograd formula for our
CG tensor product \textit{backward} kernel 
(i.e. we must provide a ``double-backward" implementation).

\newcommand{\backward}{\textrm{backward}}
To avoid spiraling engineering complexity, we
will implement the double-backward pass
by linearly combining the outputs from existing
kernels. Let $\bm z = \tp(\bm x, \bm y, \bm W)$ (we omit the
sparse tensor argument $\scr P$ here) and define 
$\bm g_z = \partial E / \partial \bm z$ for the scalar
energy prediction $E$. Finally, let
$\bm a$, $\bm b$, and
$\bm C$ be the gradients calculated by the
backward pass, given as
\begin{equation*}
\begin{aligned}
\br{\bm a, \bm b, \bm C}
&= \br{\partial E / \partial \bm x,
\partial E / \partial \bm y,
\partial E / \partial \bm W} \\
&= \textrm{backward}\paren{\bm x, \bm y, \bm W, \bm g_z}.
\end{aligned}
\end{equation*}
Now our task is to compute 
($\partial \scr L / \partial \bm x$, $\partial \scr L / \partial \bm y$,
$\partial \scr L / \partial \bm W$, $\partial \scr L / \partial \bm g_z$) given ($\partial \scr L / \partial \bm a$,
$\partial \scr L / \partial \bm b$, $\partial \scr L / \partial \bm C$). We dispatch seven calls to the
forward and backward pass kernels:
\begin{equation}
\begin{aligned}
\textrm{op1} &= \backward(\partial \scr L / \partial \bm a, 
\partial \scr L / \partial \bm b, \bm W, \bm g_z) \\
\textrm{op2} &= \backward(\bm x, \bm y, \partial \scr L / 
\partial \bm C, \bm g_z) \\
\textrm{op3} &= \tp(\partial \scr L / \partial \bm a, \bm y, \bm W) \\
\textrm{op4} &= \backward(\partial \scr L / \partial \bm a, \bm y, \bm W, \bm g_z) \\
\textrm{op5} &= \backward(\bm x, \partial \scr L / \partial \bm b, \bm W, \bm g_z) \\
\textrm{op6} &= \tp(\bm x, \partial \scr L / \partial \bm b, \bm W) \\
\textrm{op7} &= \tp(\bm x, \bm y, \partial \scr L / \partial \bm C).
\end{aligned}
\label{eq:kernel_dispatches}
\end{equation}
By repeatedly applying the 
product and chain rules to the formulas 
for $\bm a$, $\bm b$, and $\bm C$ in 
Section \ref{sec:backward_pass}, 
we can show 
\begin{equation}
\begin{aligned}
\partial \scr L / \partial \bm x &= \textrm{op1}\br{1}
+ \textrm{op2}\br{1} \\
\partial \scr L / \partial \bm y &= \textrm{op1}\br{2}
+ \textrm{op2}\br{2} \\
\partial \scr L / \partial \bm W &= \textrm{op4}\br{3}
+ \textrm{op5}\br{3} \\
\partial \scr L / \partial \bm g_z &= \textrm{op3}
+ \textrm{op6} + \textrm{op7},
\end{aligned}
\label{eq:higher_deriv_formulas}
\end{equation}
where $\textrm{op1}\br{1}$, $\textrm{op1}\br{2}$,
and $\textrm{op1}\br{3}$ denote the three
results calculated by the backward function, 
and likewise for $\textrm{op2}$, $\textrm{op4}$, and $\textrm{op5}$. 
Equations \eqref{eq:kernel_dispatches} and
\eqref{eq:higher_deriv_formulas} could be implemented
in less than 10 lines of Python and 
accelerate the double-backward pass without
any additional kernel engineering. In practice, we fuse 
the forward calls into a single kernel by calling
Algorithm \ref{alg:kernel_bc} three times with different
arguments in a procedure like 
Algorithm \ref{alg:cg_template}. The 
backward calls fuse in a similar
manner, and we adopt this approach to dramatically
reduce memory traffic and kernel launch overhead.


\subsection{Graph Convolution and Kernel Fusion}
\label{sec:graph_convolution}
Figure \ref{fig:cg_tp_applications} illustrates two typical 
use cases of the CG tensor product kernel. The first case (\ref{fig:cg_tp_applications}A)
calls the kernel illustrated in Figure \ref{fig:cg_tensor_product}
several times with unique triples 
of $(\bm x, \bm y, \bm W)$ inputs, and we have
already addressed its implementation. The 
second case (\ref{fig:cg_tp_applications}B) embeds the CG tensor product into
a graph convolution operation \cite{thomas_tensor_2018, batzner_2022,Batatia2022mace}. Here, the nodes of a graph typically 
correspond to atoms in a simulation and 
edges represent pairwise interactions. For a graph 
$G = (V, E)$, let 
$\bm x_1...\bm x_{\abs{V}}$
$\bm y_1...\bm y_{\abs{E}}$, and
$\bm W_1... \bm W_{\abs{E}}$ 
be node embeddings, edge embeddings,
and trainable edge weights, respectively. Then
each row $\bm z_j$ of the graph convolution output, $j \in \br{\abs{V}}$, is given by
\begin{equation}
\bm z_j = \sum_{(j, k, e) \in \scr N(j)} \tp (\scr P, \bm x_k, \bm y_e, \bm W_e),
\label{eq:graph_conv}
\end{equation}
where $\scr N(j)$ denotes the neighbor set of
node $j$ and $(j, k, e) \in \scr N(j)$ 
indicates that edge $e$ connects nodes $j$ and $k$.
Current equivariant message passing networks
\cite{batzner_2022, Batatia2022mace} implement
Equation \eqref{eq:graph_conv} by 
duplicating the node
features to form 
$\bm x_1', ..., \bm x_{\abs{E}}'$,
calling the large batch kernel developed earlier,
and then executing a scatter-sum (also called
reduce-by-key) to perform aggregation. 
Unfortunately, duplicating the node features 
incurs significant memory and 
communication-bandwidth overhead when 
$\abs{E} \gg \abs{V}$ (see 
Table \ref{tab:molecular_graphs}). 

Notice that
graph convolution exhibits a memory access pattern similar to 
sparse-dense matrix multiplication 
(SpMM) \cite{yang_spmm_principles}. 
We provide two procedures for
the fused CGTP / graph convolution based
on classic SpMM methods.
The first, detailed in Algorithm
\ref{alg:convolution_fusion}, requires
row-major sorted edge indices and iterates
over the phases of the computation
schedule as the outer loop. The latter
change enables the algorithm 
to keep a running
buffer $\bm z_{\textrm{acc}}$ that
accumulates the summation in
Equation \eqref{eq:graph_conv} for each node. The
buffer $\bm z_{\textrm{acc}}$ is 
only flushed to global memory when
a warp transitions to a new row
of the graph adjacency matrix, reducing
global memory writes from $O(\abs{E})$
to $O(\abs{V})$. To handle the case
where two or more warps calculate 
contributions to the same node, we write the
first row processed by each warp to
a fixup buffer \cite{yang_spmm_principles}. We 
developed a backward pass kernel
using a similar SpMM strategy, but a 
permutation that transposes the graph 
adjacency matrix is required as part
of the input.

\begin{algorithm}
\caption{Deterministic TP + Graph
Convolution}
\begin{algorithmic}
\Require Graph $G = (V, E)$, 
$E\br{b} = (i_b, j_b)$ 

\Require Edges in $E$ sorted by
first coordinate.

\Require Batch $\bm x_1,...,\bm x_{\abs{V}}$, $\bm y_1,...\bm y_{\abs{E}}$,$\bm W_1,...,\bm W_{\abs{E}}$

\For{$\textrm{seg}_i \in \textrm{schedule}$}
    \State $(s, t) = E\br{k}\br{0}, E\br{k}\br{1}$
    \State Set $\bm z_{\textrm{acc}} = 0$
    \For{$b=1...\abs{E}$}
    \Comment{Parallel over warps}
            \State $\bm x_{\textrm{smem}} = 
        \bm x_t\br{\textrm{seg}_{i\ \textrm{(x start)}}:
        \textrm{seg}_{i\ \textrm{(x end)}}}$
        \State Load $\bm y_{\textrm{smem}}, \bm W_{\textrm{smem}}$ similarly, set $\bm z_{\textrm{smem}} = 0$.

        \State Run kernels
        as in Algorithm \ref{alg:cg_template}.

        \State $\bm z_{\textrm{acc}} \pluseq \bm z_{\textrm{smem}}$

        \If{$b = \abs{E}$ or
        $s < E\br{b+1}\br{0}$}
        \If{$s$ is first
        vertex processed by warp}
        \State Send 
        $\bm z_{\textrm{acc}}$ to fixup
        buffer.
        \Else
        \State $\bm z_s\br{\textrm{seg}_{i\ \textrm{(z start)}}:
        \textrm{seg}_{i\ \textrm{(z end)}}} \pluseq \bm z_{\textrm{acc}}$
        \EndIf
        \State $\bm z_{\textrm{acc}} = 0$
        \EndIf
    \EndFor
\EndFor
\State Execute fixup kernel.
\end{algorithmic}
\label{alg:convolution_fusion}
\end{algorithm}

The second algorithm, which we
omit for brevity, functions
almost identically to Algorithm
\ref{alg:convolution_fusion}, but
replaces the fixup / store logic
with an atomic accumulation at 
every inner loop iteration.
This \textit{nondeterministic} method performs $O(\abs{E})$
atomic storebacks, but does not
require a sorted input graph or
adjacency transpose permutation.



\section{Experiments}
\label{sec:experiments}
Our kernel generator is available online\footnote{
\url{https://github.com/PASSIONLab/OpenEquivariance}
} as an installable Python package.
We adopted the frontend interface
of e3nn \cite{e3nn_paper, e3nn_software} and used QuTiP
\cite{qutip2, qutip5} to generate CG 
coefficients. We
tested correctness against e3nn to ensure that our kernels
produce identical results, up to floating
point roundoff and 
a well-defined permutation of the weights
$\bm W$ on certain input configurations. In cases where weight
reordering is required, we provide a function for easy migration.
We use the NVIDIA and AMD HIP Runtime Compilers to compile our
generated kernels through a C++ extension to Python.

\begin{table}[ht]
\centering
\begin{tabular}{ll}
\toprule
Quantity                & Value       \\
\midrule
FP32 Peak               & 19.5 TFLOP/s \\
FP64 SIMT Peak          & 9.7 TFLOP/s  \\
FP64 Tensor Core Peak   & 19.5 TFLOP/s \\
HBM2 Bandwidth & 2.04 TB/s \\
\bottomrule
\end{tabular}
\caption{A100-SXM4-80GB performance 
\cite{a100_architecture}.}
\label{tab:a100_info}
\end{table}

The majority of experiments were conducted on NVIDIA A100 GPU nodes of NERSC Perlmutter
(each equipped with an AMD EPYC 7763 CPU). Table
\ref{tab:a100_info} lists the advertised maximum 
memory bandwidth and compute peaks for 
multiple datatypes, a yardstick for our results. 
Section \ref{sec:cross_gpu_comparison} covers 
performance on other GPU models. 

\begin{figure}
    \centering 
    \includegraphics[width=1.0\linewidth]{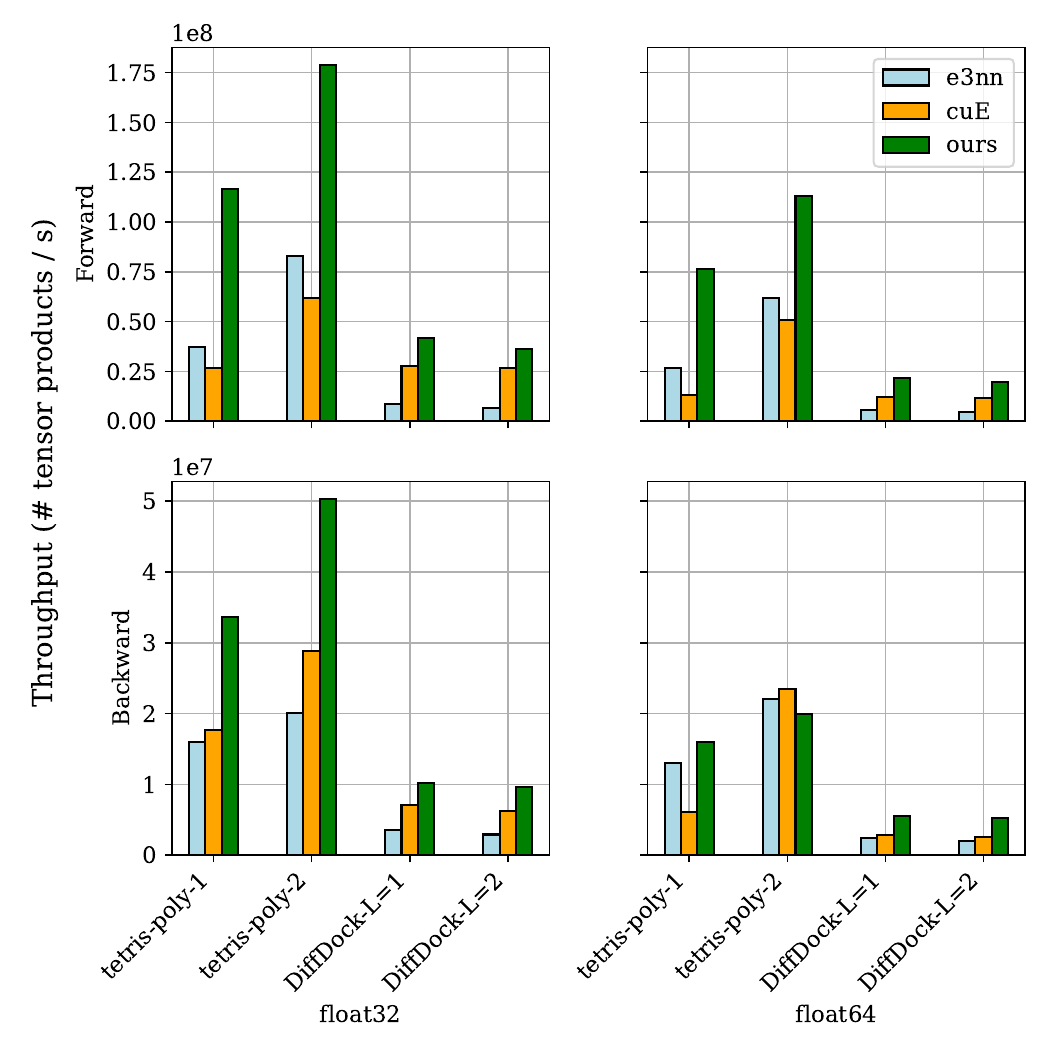}
    \caption{
    Throughput of CG tensor products (batch size 50K), kernel C configurations. We exhibit 
    up to 2x speedup over cuE on the
    DiffDock tensor products.
    } 
    \label{fig:uvw_throughput_comparison}
\end{figure}

As baselines, we used the PyTorch versions of 
e3nn (v0.5.6) \cite{e3nn_software} and NVIDIA cuEquivariance (v0.4.0)
\cite{geiger_accelerate_2024}. 
The e3nn implementation was
accelerated with \texttt{torch.compile} except
where prohibited by memory constraints. 
For Figures \ref{fig:uvu_throughput_comparison}, 
\ref{fig:uvw_throughput_comparison}, 
\ref{fig:double_backward}, and \ref{fig:kernelfusion}, we 
benchmarked all functions through a 
uniform PyTorch interface and included any 
overhead in the measured runtime. 
Figures \ref{fig:roofline_analysis} and
\ref{fig:mace-sim} (right) rely on 
kernel runtime measurements without 
PyTorch overhead.

cuEquivariance experienced a
significant efficiency increase since v0.2.0,
the latest version available when our first 
preprint was released (see Figures \ref{fig:roofline_analysis}, \ref{fig:mace-sim}). Since that early
release, the authors also added JIT capability and 
fused convolution, although
the closed source kernel backend 
renders the details opaque. Unless
otherwise noted, we report all 
benchmarks against cuE v0.4.0.

\subsection{Forward / Backward Throughput}
\begin{figure}
    \centering
    \includegraphics[width=1.0\linewidth]{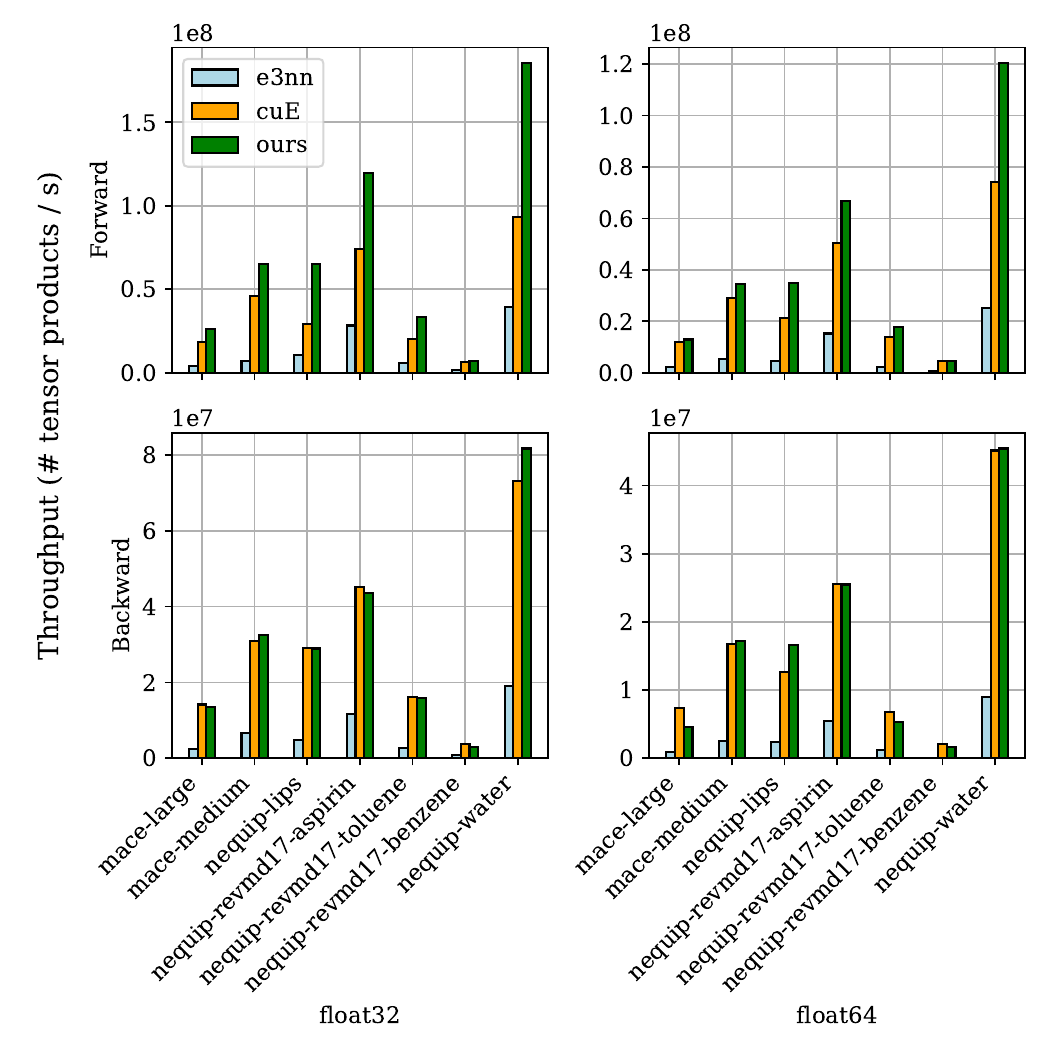}
    \caption{Throughput of CG tensor products (batch size 50K), 
    kernel B configurations without SpMM
    kernel fusion. On difficult configurations like 
    Nequip-benzene with massive output vector lengths, we exhibit more than 10x improvement over e3nn.} 
    
    \label{fig:uvu_throughput_comparison}
\end{figure}

We first profiled our kernels on a large collection of model 
configurations used by Nequip 
\cite{batzner_2022} and MACE \cite{Batatia2022mace}.
For each model, we selected an expensive representative 
tensor product to benchmark. Figure \ref{fig:uvu_throughput_comparison} shows the results of
our profiling on configurations that use only Kernel B (see 
Figure \ref{fig:cg-primitives}). 
Our median FP32 speedup 
over e3nn was 5.9x, resp. 4.8x, for 
the forward and backward passes, with a maximum
of 9.2x for the forward pass. We observed 
a median speedup of 1.6x over cuE for 
the FP32 forward pass, which drops to 1.3x 
in FP64 precision. Our median performance 
approaches parity with cuE 
on the backward pass, 
with a minimum and maximum speedup of 0.72x 
and 1.32x in FP64 precision.

To benchmark kernel C, we used the Tetris polynomial
from e3nn's documentation \cite{e3nn_software} 
and two configurations based on DiffDock 
\cite{corso2023diffdock}. We exhibit between 
1.4x and 2.0x speedup over cuE for both forward
and backward passes on DiffDock. 
Our speedups over e3nn are less 
dramatic for kernel C, which has a workload dominated by the 
small dense matrix multiplication in 
Algorithms \ref{alg:kernel_bc} and \ref{alg:kernel_b_backward}.

\subsection{Second Derivative Performance}
\label{sec:double_backward}
\begin{figure}
    \centering 
    \includegraphics[width=1.0\linewidth]{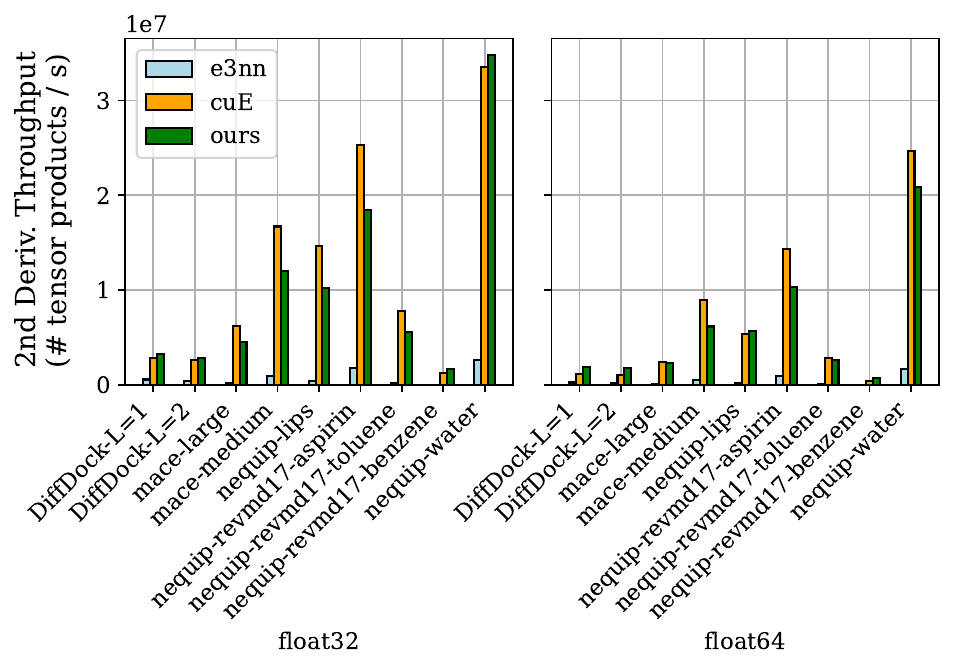}
    \caption{
    Throughput of second derivative kernels 
    (batch size 20K) for chemistry / protein models.} 
    \label{fig:double_backward}
\end{figure}

We analyzed the double 
backward pass for the tensor products 
from the prior section
(excluding the Tetris polynomial, which does not require
it). Figure \ref{fig:double_backward} shows our results. Across datatypes and
model configurations, our
speedup ranges from 5.5x to 35x over
e3nn and 0.69x-1.69x over
cuE. Although our median speedup over
cuE is 0.73x in FP32 precision and 
0.93x for FP64 precision, we exhibit 
lower runtime on all 
DiffDock tensor products and several
Nequip configurations. 

For many Nequip / MACE 
configurations, the performance gap between our
implementation and cuE
could likely shrink with some judicious
kernel tuning. In particular, we could improve our heuristic selection of 
the warp count per block, the 
number of blocks, and the shared
memory allotted to each block. We leave
tuning these hyperparameters as future work. 

\subsection{Roofline Analysis} 
We conducted a roofline analysis \cite{roofline} by profiling our 
forward / backward 
pass implementations 
on varied input configurations.
We profiled tensor products
with a single ``B" subkernel 
(see Figure \ref{fig:cg-primitives})
with FP32 precision, 
core building blocks for
models like Nequip and MACE. The 
arithmetic intensity of the CG
tensor product depends on the 
structure of the sparse tensor, and we profiled configurations
with progressively increasing arithmetic intensity.

\begin{figure}
    \centering
      \includegraphics[width=1.0\linewidth]{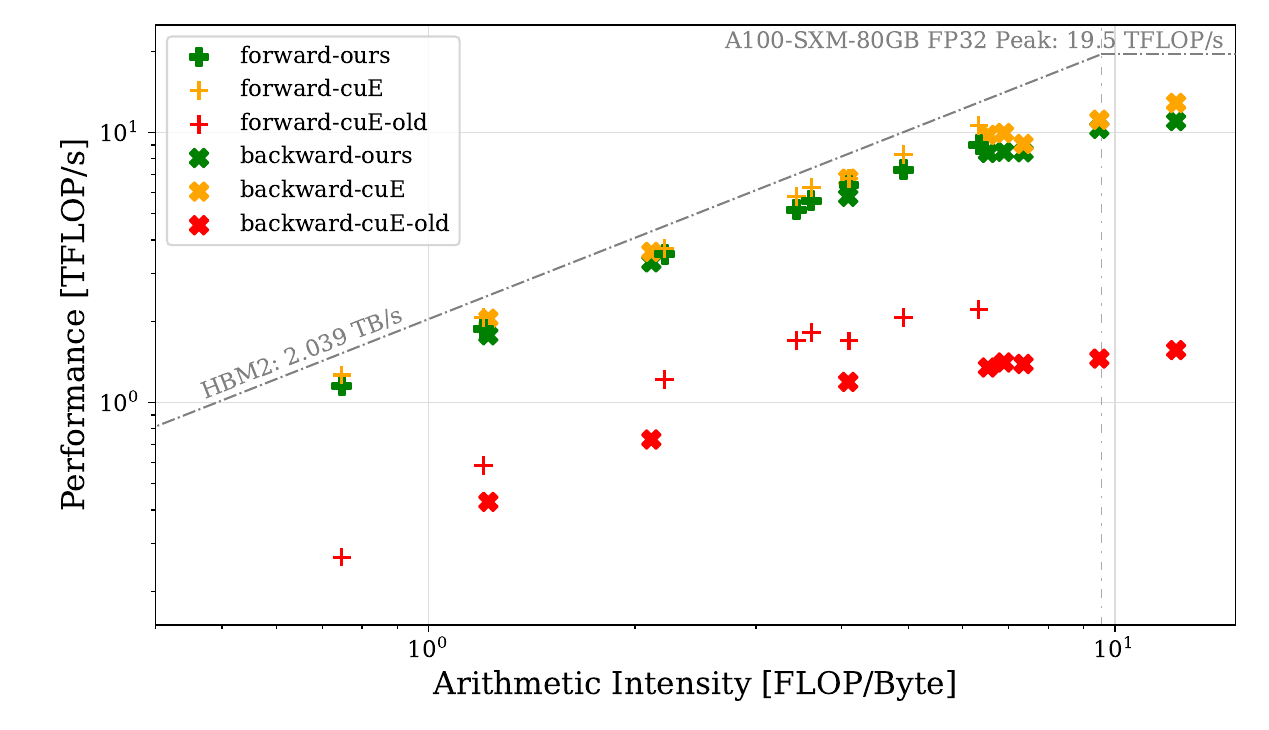}
    \caption{Roofline analysis for input
    configurations of varying arithmetic intensity, batch size 200K. Our
    kernels and the latest version of
    cuE closely track the slope
    of the global memory roofline, indicating high
    efficiency. An older version of cuE (v0.2.0) is also
    included to highlight the 
    performance improvement in their package 
    (see top of Section \ref{sec:experiments}).} 
    \label{fig:roofline_analysis}
\end{figure}

Figure \ref{fig:roofline_analysis} shows 
our results, 
which indicate consistently high compute and bandwidth utilization for both our kernels and the latest version
of cuE. An earlier version 
of cuE (v0.2.0, indicated
on the plot as cuE-old) exhibited significantly lower 
efficiency, which has since been corrected by the package
authors. The performance of all kernels
saturates at 58\% of the FP32 peak, likely because 
Algorithms \ref{alg:kernel_bc} and 
\ref{alg:kernel_b_backward} contain contain a significant fraction
of non fused-multiply-add (FMA) instructions.

\subsection{Additional GPU Models}
\label{sec:cross_gpu_comparison}
We also tested our kernels on the NVIDIA
A5000 and a single GPU die of the AMD 
MI250x. Table \ref{tab:cross_gpu} compares the
MACE tensor product runtime 
across architectures and kernel providers; 
our codebase contains a more 
complete set of benchmarks. The A5000 performance matches our expectations given
its lower memory bandwidth compared to the A100. 
While we also saw significant speedup on the MI250x,
we detected somewhat lower memory bandwidth utilization 
than predicted. 

\begin{table}[ht]
\centering
\begin{tabular}{lllllll}
\toprule
\multirow{2}{*}{GPU}    & \multicolumn{3}{c}{forward} & \multicolumn{3}{c}{backward} \\
\cmidrule{2-7} 
       & e3nn     & cuE    & ours    & e3nn     & cuE     & ours    \\
\midrule
A100   & 13       & 2.8    & \bf{2.0}     & 21       & \bf{3.5}     & 3.7     \\
A5000  & 29       & 4.2    & \bf{3.8}     & 42       & \bf{9.3}    &  11       \\
MI250x & 41       & -      & \bf{3.0}     & 128      &   -     & \bf{15}
\\
\bottomrule
\end{tabular}
\caption{MACE-large isolated 
tensor product runtime (ms), batch size 50K, FP32 unfused.}
\label{tab:cross_gpu}
\end{table}

\subsection{Kernel Fusion Benchmarks}
\label{sec:kernel_fusion}
We conducted our kernel fusion experiments on
three molecular structure graphs listed in
Table \ref{tab:molecular_graphs}. We downloaded
the atomic structures of human 
dihydrofolate deductase
(DHFR) and the SARS-COV-2 glycoprotein spike
from the Protein Data Bank and constructed a
radius-neighbors graph for each using Scikit-Learn
\cite{scikit-learn}. The carbon lattice 
was provided to us as a representative workload 
for MACE \cite{batatia2024foundationmodel}.

\begin{table}[ht]
\centering
\begin{tabular}{lcc}
\toprule
Graph & Nodes & Adj. Matrix NNZ\\
\midrule
DHFR 1DRF & 1.8K & 56K \\
COVID spike 6VXX & 23K & 136K \\
Carbon lattice & 1K & 158K \\
\bottomrule
\end{tabular}
\vspace{8pt}
\caption{Molecular graphs for kernel fusion
experiments.}
\label{tab:molecular_graphs}
\end{table}

\begin{figure}
    \centering
    \includegraphics[width=0.85\linewidth]{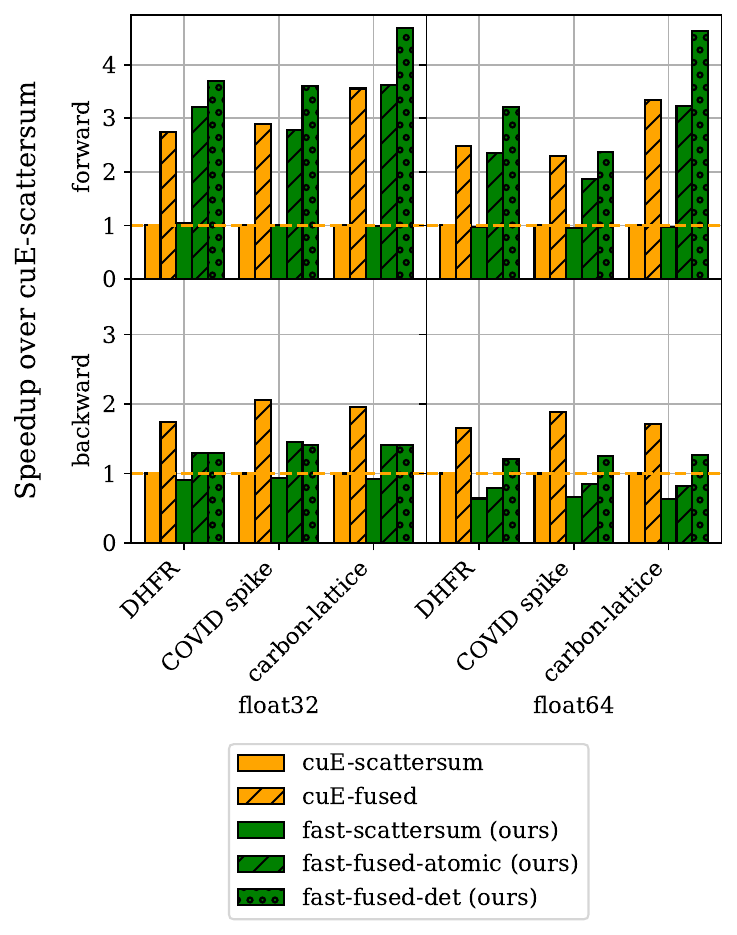}
    \caption{Speedup of convolution kernels 
    over cuE-scattersum, which
    calls cuEquivariance and follows it by a
    scatter-sum operation. cuE-fused refers to
    a new kernel introduced in cuE v0.4.0.}
    \label{fig:kernelfusion}
\end{figure}

Figure \ref{fig:kernelfusion} shows the speedup 
of fused implementations benchmarked on the most
expensive tensor product in the MACE-large model.
The baseline, ``cuE-scattersum", implements the
unfused strategy in 
Section \ref{sec:graph_convolution} by duplicating
node embeddings, executing a large batch of
tensor products with cuEquivariance, and finally performing row-based 
reduction by keys.
Our deterministic fused algorithm offers
the greatest speedup in FP64 precision on the carbon
lattice forward pass over cuE (roughly 1.3x speedup).
On the other hand, cuE-fused offers 1.3-1.4x speedup
on the backward pass, and we aim to
close this performance gap.

\subsection{Acceleration of Nequip and MACE}
The Nequip 
\cite{batzner_2022, nequip_updated_software}
and MACE 
\cite{Batatia2022mace, batatia2024foundationmodel} 
interatomic potential 
models implement the equivariant graph
neural network architecture in Figure
\ref{fig:cg_tp_applications}B. Both
have similar message passing structures,
while MACE incorporates higher order
interactions for its node features.
Our first benchmark uses the Nequip-ASE 
calculator interface
to evaluate forces on a large box of
water molecules. Due to recent 
updates to Nequip's software 
\cite{nequip_updated_software}
and time constraints, we only used the 
nondeterministic fused convolution for our 
measurements, which appear in Table
\ref{tab:nequip_speedup}. For simplicity, we 
report speedup with respect
to the Nequip Python interface 
without JITScript or \texttt{torch.compile}, 
although our package
is fully compatible with these subsystems.

\begin{table}[ht]
\centering
\begin{tabular}{@{}lll@{}}
\toprule
\multirow{2}{*}{GPU} & \multicolumn{2}{l}{Speedup over Unmodified Nequip} \\ \cmidrule(l){2-3} 
                     & ours-unfused          & ours-fused          \\ \midrule
A100                 & 6.3x                 & 7.8x               \\
MI250x               & 3.9x                 & 4.4x               \\ \bottomrule
\end{tabular}
\vspace{8pt}
\caption{Force evaluation speedup for our kernels on a 
4-layer FP32 Nequip model, 5184-atom 
water box system.}
\label{tab:nequip_speedup}
\end{table}

For MACE, we patched the code to 
sort nonzeros of 
the atomic adjacency matrix according to 
compressed sparse row (CSR) order.
We then substituted our
deterministic fused convolution into 
the model and conducted our benchmark
on the carbon lattice in 
Table \ref{tab:molecular_graphs}. 
MACE uses a distinct set of weight matrices
for each atomic species, and an 
inefficiency in the baseline 
code causes its runtime to increase
disproportionately to the useful computation
involved. Our model has a species 
dictionary of eight elements (to trigger 
the problem in the baseline code, even though
the carbon lattice only requires one), and
our package includes a module to optimize
away the inefficiency.

\begin{figure}
    \centering
    \includegraphics[width=0.89\linewidth]{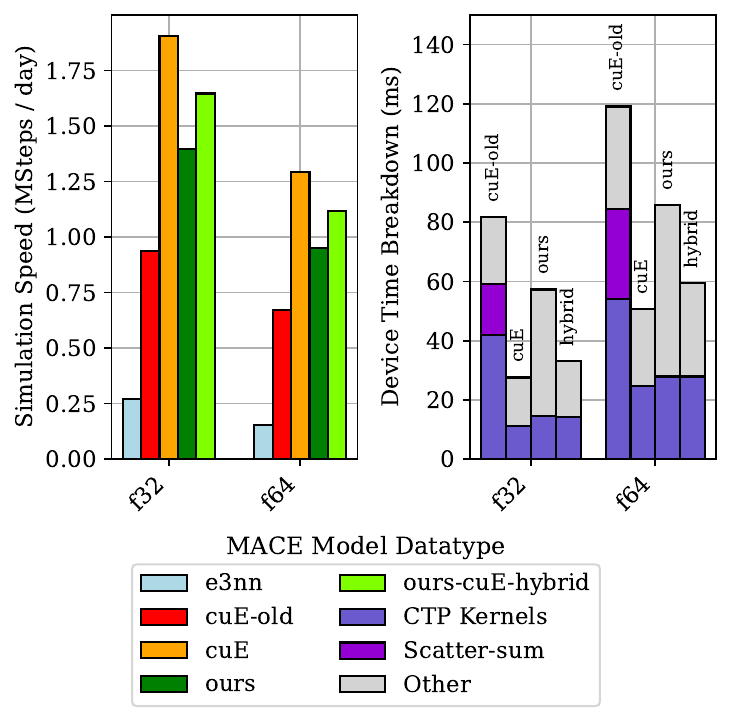}
    \caption{Simulation speed of MACE for varying
    kernel provider (left) and device time breakdown (right). cuE-old refers to
    version 0.2.0 of their package, while
    the hybrid implementation combines 
    our fused convolution with other model
    primitives optimized by cuE.}
    \label{fig:mace-sim}
\end{figure}

Figure \ref{fig:mace-sim} (left) compares the rate of 
molecular dynamics simulation among the different
kernel providers. We benchmarked cuE with the
optimal data layout for its irreps and included optimizations
for symmetric tensor contraction, linear
combination layers, and graph convolution.
In FP32 precision, we provide a 5.1x speeedup
over e3nn and 1.5x over the older 
v0.2.0 cuE package, noting that the
latter does not provide kernel
fusion. A similar speedup
exists for the FP64 models.
Our implementation, however, achieves
0.73x speedup compared to the 
latest version of cuE that introduces
kernel fusion.

To further investigate these benchmarks,
Figure \ref{fig:mace-sim} (right) breaks down device runtime 
spent in various kernels. 
Our time spent on the tensor product 
(CTP kernels) falls within 
2-3 milliseconds of cuE, a highly
competitive result. Our performance
suffers due to the
remaining model components, which
contribute to less than 15\% of the 
unoptimized model runtime. To 
address this, we 
created a hybrid model (ours-cuE-hybrid)
that combines our fused convolution with
the linear / symmetric contraction layers
offered by cuE. While the hybrid model
closes the gap further, the runtime 
of the remaining kernels is still higher
than cuE. This is because the hybrid
model preserves the original data layout
of MACE layers, whereas cuE transposes
several key weight matrices to achieve 
higher performance. As a consequence,
cuE requires a data reordering function 
for models trained without the package, 
whereas our kernels have no such restriction.

\section{Conclusions and Further Work}
\label{sec:conclusions}
We have established that our sparse kernels achieve 
consistently high performance on several
common primitives
used in O(3)-equivariant deep neural networks.  We see several avenues for future progress:

\begin{itemize}
\item \textbf{Low Precision Data and MMA Support}: 
Our kernels rely on the single-instruction multiple
thread (SIMT) cores for FP32 and FP64 floating
point arithmetic. Modern GPUs offer specialized
hardware for lower-precision calculation, both
using SIMT cores and within 
matrix-multiply-accumulate (MMA) units.
We hope to harness these capabilities in the future.

\item \textbf{Stable Summation During Convolution}: Our 
kernel generator allows us to easily extend our methods 
to use stable (Kahan) summation \cite{kahan_summation} 
within fused graph convolution. Kahan summation 
reduces numerical roundoff error during feature vector aggregation
across node neighborhoods, promoting energy 
conservation in simulations. 

\item \textbf{Integration into new models}: 
Our open-source software remains accessible to 
newcomers while delivering the high performance required for massive workloads. In conjunction with domain experts, we hope to apply our library
to train larger, more expressive equivariant
deep neural networks. 
\end{itemize}
\newpage

\section*{Acknowledgements and Disclaimers}
We thank the anonymous referees for feedback that 
strengthened our draft. This research was supported by the 
U.S. Department of Energy, Office of Science, Office of 
Advanced Scientific Computing Research, Department of
Energy Computational Science Graduate Fellowship under Award Number 
DE-SC0022158. This research is also supported by the Office of Science 
of the DOE under Award Number DE-AC02-05CH11231. We used resources 
of the National Energy Research Scientific Computing Center (NERSC), 
a Department of Energy Office of Science User Facility 
using NERSC award ASCR-ERCAP-33069. We also used resources of the Oak Ridge Leadership Computing Facility at the Oak Ridge National Laboratory, which is supported by the Office of Science of the U.S. Department of Energy under Contract No. DE-AC05-00OR22725.

This report was prepared as an account of work sponsored by an agency of the
United States Government. Neither the United States Government nor any agency thereof, nor
any of their employees, makes any warranty, express or implied, or assumes any legal liability
or responsibility for the accuracy, completeness, or usefulness of any information, apparatus,
product, or process disclosed, or represents that its use would not infringe privately owned
rights. Reference herein to any specific commercial product, process, or service by trade name,
trademark, manufacturer, or otherwise does not necessarily constitute or imply its
endorsement, recommendation, or favoring by the United States Government or any agency
thereof. The views and opinions of authors expressed herein do not necessarily state or reflect
those of the United States Government or any agency thereof.

\bibliographystyle{plainnat}
\bibliography{bibliography}

\end{document}